\documentclass[runningheads]{llncs}
\usepackage{graphicx}
\usepackage[colorlinks,
            linkcolor=red,
            anchorcolor=red,
            citecolor=green]{hyperref}
            
\usepackage{booktabs}  
\usepackage{threeparttable} 
\usepackage{multirow}
\usepackage{wrapfig}
\usepackage{bbding} 

\usepackage{tikz}
\usepackage{comment}
\usepackage{amsmath,amssymb} 
\usepackage{color}

\usepackage[accsupp]{axessibility}  

\begin{document}
\pagestyle{headings}
\mainmatter
\def\ECCVSubNumber{3084}  

\title{Long-Tailed Classification with Gradual Balanced Loss and Adaptive Feature Generation} 



\titlerunning{Abbreviated paper title}
%
\author{Zihan Zhang \and Xiang Xiang}
\authorrunning{Z. Zhang and X. Xiang}
%
\institute{School of Artificial Intelligence and Automation,\\
Huazhong University of Science and Technology, China\\
\email{xex@hust.edu.cn}
}

\maketitle

\begin{abstract}
The real-world data distribution is essentially long-tailed, which poses great challenge to the deep model. In this work, we propose a new method, Gradual Balanced Loss and Adaptive Feature Generator (GLAG) to alleviate imbalance. GLAG first learns a balanced and robust feature model with Gradual Balanced Loss, then fixes the feature model and augments the under-represented tail classes on the feature level with the knowledge from well-represented head classes. And the generated samples are mixed up with real training samples during training epochs. Gradual Balanced Loss is a general loss and it can combine with different decoupled training methods to improve the original performance. State-of-the-art results have been achieved on long-tail datasets such as CIFAR100-LT, ImageNetLT and iNaturalist, which demonstrates the effectiveness of GLAG for long-tailed visual recognition.


\end{abstract}

\section{Introduction}
\begin{figure}
\centering
\includegraphics[scale=0.23]{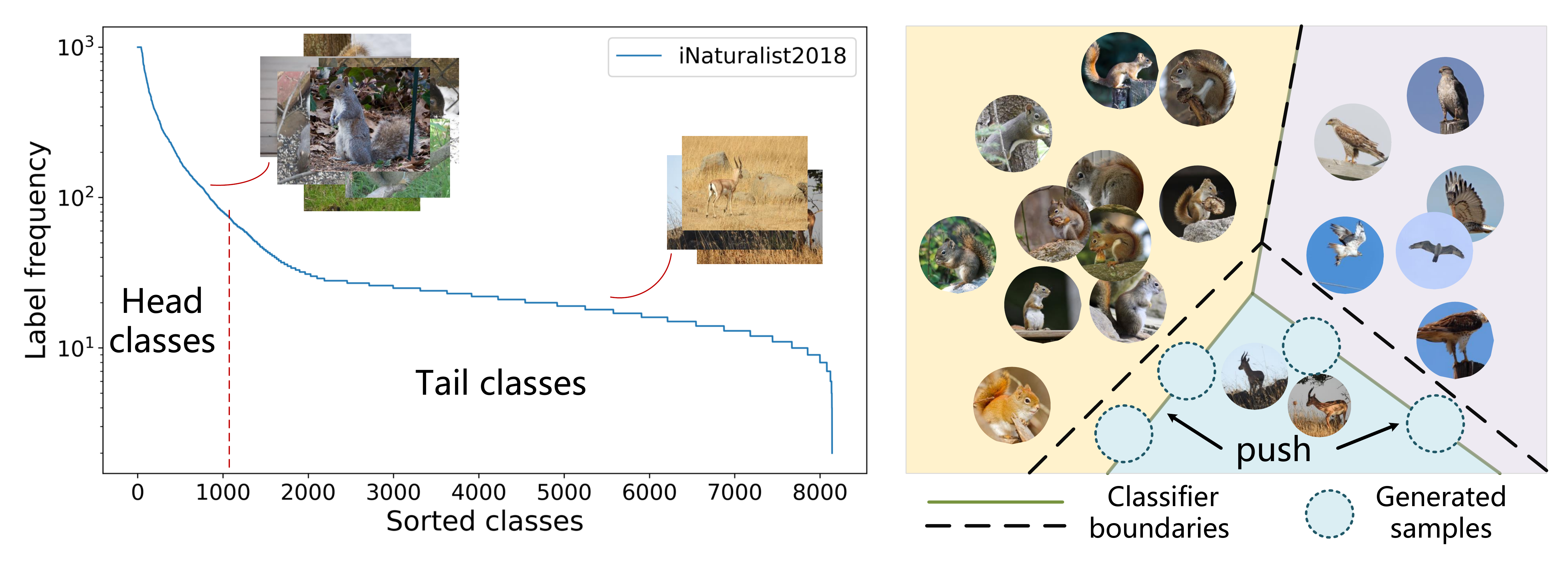}
\vspace{-3mm}
\caption{(a): Visualization of label frequency of the real-world dataset iNaturalist 2018. (b): Illustration of data augmentation method for long-tail problem aiming to recover the underlying distribution.}
\label{fig:Introduction}
\end{figure}

While deep neural networks have shown incredible success in image classification tasks \cite{he2015delving}, it is primarily driven by large-scale data, e.g., CIFAR and ImageNet \cite{deng2009imagenet}. Such datasets are balanced, where different classes have approximately equal number of instances. However, the real-world is naturally long-tailed, that is, a few classes (\emph{i.e.}, head classes) contain a large number of samples while most classes (\emph{i.e.}, tail classes) contain few samples \cite{japkowicz2002class,zhang2021deep}. This is easily illustrated through examples we encounter in life: we often see dogs and cats in city, but rarely see tigers or wolves. That is also illustrated by real-world datasets such as LVIS \cite{gupta2019lvis} and iNaturalist \cite{van2018inaturalist} which is shown in Fig.~\ref{fig:Introduction}(a). 

When dealing with long-tail datasets, the performance of deep neural networks with standard training suffers much \cite{liu2019large,wang2019dynamic}. The task is challenging due to two difficulties: (1) lack of tail class samples makes it difficult to train the feature model that generalizes well in tail classes, which is also the challenge in few-shot learning \cite{yang2021free,dhillon2019baseline,cao2019theoretical}. (2) the dominance of head classes makes the classifier biased to the head classes \cite{kang2019decoupling}.

An active strand of work to tackle the long-tail challenge is class re-balancing including data re-sampling \cite{cai2021ace,wang2020devil}, re-weighting loss learning \cite{cui2019class,wang2021seesaw}, and logit adjustment \cite{menon2020long}. Re-sampling is widely-used to alleviate imbalance by over-sampling the tail classes and under-sampling the head classes \cite{chawla2002smote,dablain2022deepsmote}. Logit adjustment aims to shift the logit to increase the margin of tail class boundary, and it can be implemented by post-hoc and the logit adjusted loss which is similar to re-weighting loss learning. Kang et al. \cite{kang2019decoupling} decouple the feature model  training and classifier training and they find the original long-tailed data without re-sampling strategy benefits the representation learning most. 

Another strand is to augment the tail classes \cite{yin2019feature,kim2020m2m,liu2020deep}. RSG \cite{wang2021rsg} transfers feature displacement from head classes to tail classes to enlarge the feature space of tail classes. M2m \cite{kim2020m2m} tries to translate the images of head classes to tail classes to balance the dataset. OFA \cite{chu2020feature} uses the class-generic information extracted by class activation maps to recover the distribution of the tail classes from the class-specific information.

In this work, we first perform an ablative analysis on the performance of feature model trained with different classification loss including cross-entropy and re-weighting loss. Specifically speaking, we first train deep models with different loss on the same long-tail CIFAR-100 dataset, then fix the feature model and retrain the classifier from the scratch on balanced full CIFAR-100 dataset. We find that, although the end-to-end performance of re-weighting loss is much higher than cross-entropy \cite{cao2019learning,menon2020long,ren2020balanced}, the feature model trained with them is not the best for decoupled training. 

To overcome their limitations, we propose a simple and yet effective modification of softmax cross-entropy: Gradual Balanced Loss (GraLoss). GraLoss can improve the performance of feature model in decoupled training, and it outperforms cross-entropy with instance-balancing strategy which is widely adopted in decoupled training. That means we can simply replace the cross-entropy loss to GraLoss in stage one to achieve a better performance. Specifically speaking, different from several recent proposals, GraLoss dynamically adjusts the logit output of classifier with a trade-off parameter $\alpha$. With GraLoss, the feature model can be more robust and perform better both in head and tail classes than previous loss function including cross-entropy and logitloss \cite{menon2020long}, which will benefit the second stage training. 

In the second stage, we focus on transferring the knowledge of head classes to tail classes to recover the distribution of tail classes as shown in Fig.~\ref{fig:Introduction}(b). To be specific, we develop an Adaptive Feature Generator (AFG) to augment the tail class samples. AFG contains a class-wise adaptive parameter $\beta$ as the confidence score of the knowledge learned from head classes. In practice, $\beta$ is adaptively changed according to the validation accuracy, which allows us to achieve a flexible and confidence-aware distribution calibration for tail classes. 

Our contributions can be summarized as follows: (1) We introduce Gradual Balanced Loss which is a general model for decoupled training that enables the feature model more balanced and robust than previous loss function in stage one; (2) We propose an effective feature generator which aims to generate the tail class samples adaptively to alleviate the bias of the classifier; (3) We evaluate our model on different long-tail image datasets, and the model outperforms previous works and achieves state-of-the-art performance on long-tail image classification.

\section{Related Work}
In this section, We first review related works on long-tail classification with our proposed approach, and then discuss the difference and relation of our approach to them.
\subsection{Re-weighting Loss Function}
Re-weighting loss function is a popular approach to tackle long-tail challenge \cite{cui2019class,cao2019learning,wu2020distribution,wang2021seesaw,tan2021equalization,ren2020balanced,menon2020long}. Cross-Entropy is widely adopted in balanced image classification. However, on long-tail dataset, the classifier trained by cross-entropy is highly biased. One of the reason is that it enhances the accuracy across the whole training samples, not on the training class level. Another reason is that gradients in the deep model brought by tail class samples are overwhelmed by gradients from head class samples \cite{wang2021seesaw}. 

Re-weighting loss function adjusts the loss values including \textit{input} value and \textit{output} value of the loss function based on the frequency of each class in the training set. Specifically, the \textit{output} value of loss function refers to the calculated loss terms. Focal loss \cite{lin2017focal} uses the prediction hardness to re-weight classes and assigns higher weights to the loss terms corresponding to the tail classes. The \textit{input} value of loss function refers to the logit output of classifier. With the small amount of samples, the tail classes often get lower prediction scores from the classifier \cite{wu2020distribution}. Therefore, an intuitive method is to directly adjust prediction scores based on the label frequencies \cite{cao2019learning,ren2020balanced,menon2020long}. To be specific, these approaches first transform label frequencies to class-wise scores, then add the inverse scores to the logit output from the classifier. That ensures the prediction scores of different classes more balanced and encourages tail classes to have larger margins. 

Our GradLoss follows the latter branch of re-weighting methods. In comparison, the weights of label frequencies used to calibrate the prediction scores in GradLoss is not fixed to a certain number. We find that, when the weights increase gradually from zero to a certain number, the performance of feature model is better than that with fixed weight and GradLoss can simply combine with other decoupled training methods to improve their performance.

\subsection{Data Augmentation}
When tackle with long-tail problem, a natural idea is to augment the tail classes to make the dataset balanced \cite{zang2021fasa,yin2019feature,kim2020m2m,liu2020deep,chu2020feature,wang2021rsg,Li_2021_CVPR,li2021self,vigneswaran2021feature}. In general, we can group related works in two categories: data augmentation \textit{with} and \textit{without} transfer learning. Data augmentation methods \textit{with} transfer learning \cite{li2021self,wang2021rsg,chu2020feature,kim2020m2m,yin2019feature,wang2021label} seek to introduce additional information for tail classes. For example, in \cite{chu2020feature}, class activation maps are used to extract class-specific features and class-generic features. The class-specific features of tail classes are then combined with class-generic features of head classes to generate new tail class samples. Inspired by \cite{yang2021free}, TailCalibration \cite{vigneswaran2021feature} tries to estimate and calibrate the feature distribution of tail classes to sample additional features for tail classes.  

Data augmentation methods \textit{without} transfer learning aim to augment the tail class samples by conventional augmentation methods such as Mixup \cite{zhang2017mixup}. MiSLAS \cite{zhong2021improving} explores using mixup augmentation method to enhance the representation learning, which improves the performance of decoupled training. FASA \cite{zang2021fasa} estimates the mean and variance of different classes during training, then samples additional data points with an adaptive sampling rate. 

Following \cite{vigneswaran2021feature,yang2021free}, we transfer the knowledge of head classes to tail classes which helps to recover the true distribution of tail classes. Different from them, we introduce an adaptive and weighted strategy to calibrate the tail classes distribution more flexibly and robustly without expensive parameter tuning.

\subsection{Decoupled training}
Decoupled training methods decouple the learning process into representation learning (\emph{i.e.}, stage one) and classifier training (\emph{i.e.}, stage two) \cite{kang2019decoupling,zhong2021improving,zhang2021distribution,chu2020feature,li2021self,xiang2021coarsetofine}. Decoupling \cite{kang2019decoupling} proposes a two-stage training method based on the finding: the model trained with end-to-end training method suffers from the imbalanced dataset. Consequently, a decoupling training strategy is proposed to address the limitations. To be specific, in stage one, the whole model is trained end-to-end with instance-balanced sampling strategy. Then, in stage two, the feature model is frozen and the classifier is trained with class balanced sampling strategy. 

Several recent studies follow the training scheme and focus on enhancing the performance of classifier. For instance, DisAlign \cite{zhang2021distribution} decouples the training scheme and adjusts the logit output with an adaptive confidence score.  

The sampling strategies in decoupled training have been fully discussed in the previous work, but how the re-weighting loss affects the feature model, not the classifier remains insufficiently discussed. That is different from previous data augmentation methods like RSG \cite{wang2021rsg} and M2m \cite{kim2020m2m} which combine the re-weighting loss method and augmentation method end-to-end. In this work, we focus on the effect of different re-weighting loss to the feature model. We want to investigate whether we can obtain a better feature model with re-weighting loss than cross-entropy, and we also want to answer the question whether the cross-entropy with instance-balancing strategy the best choice to train a feature model in decoupled training or not.

\section{Problem Setting and Ablative Analysis}
In this section, we first introduce the setting of long-tail classification task and the decoupled training scheme, then we analyze the influence of different re-weighting loss to the feature model in decoupled training. 

\subsection{Problem Setting}
Given an long-tail dataset $D=\{{(x_i,y_i)\}}_{i=1}^{N}$ with $N$ training samples and $L$ classes, $x_i$ is the training image and $y_i \in \{1,2,...,L\}$ is its label. For different sub-set of $D$, we denote it as $D_k$ belonging to category $k$. Defining the number of samples of $D_k$ to $N_k=|D_k|$, we have $N_1\ge N_2\ge ...\ge N_L$ and $N_1\gg N_L$ after sorting of $N_k$. The task of long-tail visual recognition is to learn a model on the long-tail training dataset $D_{train}$ that generalizes well on a balanced test split $D_{test}$.

For decoupled training, we denote $M(x_i;\theta)=\hat{y_i}$ as the classification model, where $x_i$ is the input, $\hat{y_i}$ is the prediction label and $\theta$ is the parameter of the model. The model $M(x_i;\theta)=\hat{y_i}$ contain two components: a feature model $f(x_i)=\widetilde{x_i}$ and a linear classifier $h(\widetilde{x_i})={z_i}$, where $\widetilde{x_i}$ denotes the feature of input $x_i$ and $z_i$ denotes the logit output of classifier. The prediction label is given as $\hat{y_i}=argmax(z_i)$. In this work, we focus on the adjustment of $z_i$ in stage one and the tuning of linear classifier $h(\cdot)$ in stage two.

In addition, the reason why the long-tail visual task is challenging: (1) The number of tail samples is scarce, which makes it difficult to train feature model $f(\cdot)$ on the long-tail train split that generalizes well on tail classes. (2) the dominance of head classes makes the classifier biased to the head classes, which is, the prediction score of head classes is much higher than that of tail classes. The two components of our proposed method (GraLoss and AFG) aim to tackle these challenges.

\subsection{Ablative Analysis} \label{Ablative Analysis}
As we explained, it is difficult to train a robust feature model on tail classes. Previous decoupled training works train the feature model with cross-entropy loss function and class-balancing sampling strategy in stage one \cite{kang2019decoupling,vigneswaran2021feature,zhang2021distribution}. And the re-weighting loss function methods often train the model end-to-end \cite{cao2019learning,menon2020long,wang2021seesaw}. While such designs achieve certain success, it is interesting to ask: would re-weighting loss function help the decoupled training? As follows, we attempt to address the question by ablative experiment on CIFAR-100 dataset. To be specific, we choose CIFAR-100 dataset with imbalance ratio 100 and train the deep model with different loss function. Then, we fix the feature model and retrain the classifier from the scratch  on the full CIFAR-100 dataset. We finally test the model on the test split. 

Under this setting, we test the feature model from the stage one whether it can generalize well on a large and balanced dataset. If the performance of the final model is great, it demonstrates that the feature model can really extract separable feature for different classes from a long-tail dataset. That separable feature is indeed what decoupled training need from stage one. Our results are shown in Table~\ref{table:Ablative Analysis}, which indicates that the cross-entropy is strong enough to train a feature model in decoupled training and it outperforms most of the re-weighting loss function. On the other hand, it is interesting that the parameter $\tau=0.5$ is not the optimal choice in \cite{menon2020long}, its performance on stage one is lower than that of $\tau=1.0$ and the end-to-end training result of $\tau=0.5$ is also lower than that of $\tau=1.0$. That means, in end-to-end training with re-weighting loss function, the classifier is extremely enhanced and performs balanced. However, the balanced and overall performance of the feature model is impaired to varying degrees.

Moreover, we find that logit adjust loss \cite{menon2020long} ($\tau=0.5$) outperforms cross-entropy slightly, which indicates that re-weighting loss function can indeed improve the feature model not just the classifier to some extent. Inspired by that, we introduce Gradual Balanced Loss (GraLoss) to better improve the feature model in stage one during decoupled training. The details will be discussed in the next section.

\setlength{\tabcolsep}{4pt}
\begin{table}
\begin{center}
\caption{Ablative analysis on feature model trained with different loss function. Stage One: the performance on the same CIFAR100-LT dataset. Stage Two: the average performance stage two on the full CIFAR-100 dataset. }
\label{table:Ablative Analysis}
\begin{tabular}{cccccc}
    \toprule
    \multirow{2}{*}{Method}&  
	\multicolumn{4}{c}{Stage One}& \multirow{2}{*}{Stage Two} \\
	\cmidrule(lr){2-5}  
	& Many & Medium & Few & Average & \\
	\hline
    Cross-Entropy  & 75.2 & 46.0 & 13.9 & 46.6 & {57.33}\\
    LDAM \cite{cao2019learning} & 71.8 & 48.9 & 17.1 & 47.4 & 57.21\\
    Balanced Softmax \cite{ren2020balanced}  & 67.2 & 51.0 & 31.4 & 50.8 & 56.74\\
    Logit Adjust \cite{menon2020long} ($\tau$=1.5) & 60.4 & 49.0 & 39.5 & 50.1 & 57.16\\
    Logit Adjust \cite{menon2020long} ($\tau$=1.0) & 67.1 & 50.9 & 31.4 & 50.7 & 56.63\\
    Logit Adjust \cite{menon2020long} ($\tau$=0.5) & 73.3 & 48.1 & 21.2 & 48.8 & 57.35\\
    \bottomrule
\end{tabular}
\end{center}
\end{table}
\setlength{\tabcolsep}{1.4pt}

\vspace{-5mm}

\section{Method}
\begin{figure}
\centering
\includegraphics[height=5.1cm]{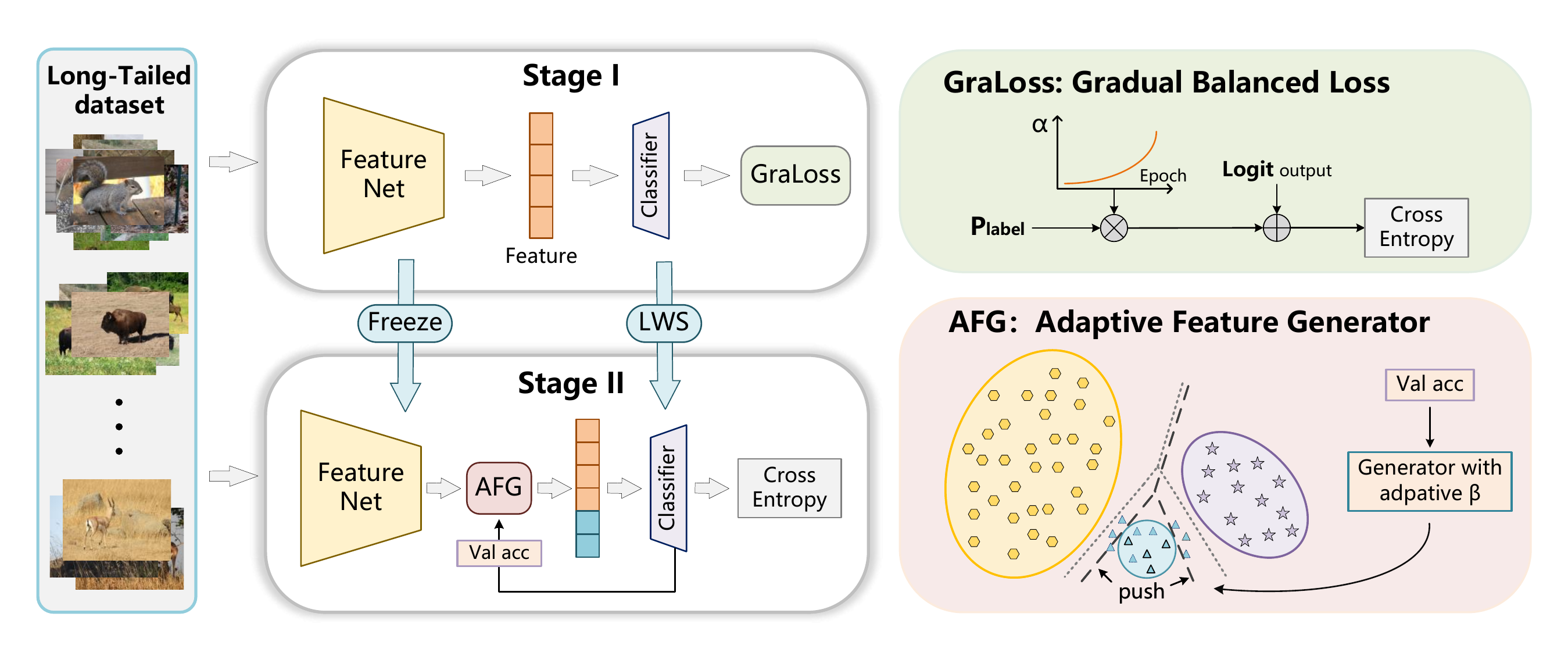}
\caption{Overview of our proposed method GLAG. Val acc: validation accuracy. LWS: Learnable weight scaling method from \cite{kang2019decoupling}.}

\label{fig:LT pipeline}
\end{figure}

In this section, we describe the proposed GLAG method in detail. Fig.~\ref{fig:LT pipeline} illustrates the overview of GLAG method.

\subsection{Gradual Balanced Loss}
 As mentioned, we define feature model as $f(x_i)=\widetilde{x_i}$ and linear classifier as $h(\widetilde{x_i})={z_i}$, where $\widetilde{x_i}$ denotes feature of input $x_i$ and $z_i$ denotes the logit output of classifier. The prediction label is given as $\hat{y_i}=argmax(z_i)$ and the label ${y_i}\in \{ 1,2,...,L\}$. For training with cross-entropy, we minimize the loss function as
 
\begin{align}
  \emph{loss} (y,z) & = -{\rm log} \frac{e^{z_y}}{\sum_{y'}^{L} e^{z_{y'}}} = {\rm log}\Big[1+\sum_{y'\neq y}^{L} e^{z_{y'}-z_y}\Big].
\end{align}
 
With the long-tail setting, the label distribution is highly skewed. The $z_y$ belonging to tail classes is rare, which results the gradient brought by the loss of head classes is much higher than that of tail classes. The re-weighting loss method tries to alleviate the imbalance on the gradient. In logit adjust \cite{menon2020long}, the logit adjusted cross-entropy loss presents as
 
\begin{align}
  \emph{loss} (y,z) & = -{\rm log} \frac{e^{z_y+\tau \cdot {\rm log}\,{p_y}}}{\sum_{y'}^{L} e^{z_{y'}+\tau \cdot {\rm log}\,{p_{y'}}}} = {\rm log} \Big[ 1+  \sum_{y'\neq y}^{L} \Big( \frac{p_{y'}}{p_{y}}\Big)^{\tau} \cdot e^{z_{y'}-z_y} \Big]\label{logit-adj loss},
\end{align}
where the ${p_y}$ is the probability distribution of label and $\tau$ is a hyper-parameter. Compared to the cross-entropy loss, logit adjusted cross-entropy loss applies a label-dependent offset to each logit. Assuming $y_t$ as the label of tail classes and $y_h$ as the label of head classes, the loss of $y_t$ is larger while the loss of $y_h$ is smaller in (\ref{logit-adj loss}). Therefore, the model has to focus more on the tail classes to decrease the loss, which will make the model balanced. 

Different from the fixing parameter in \cite{menon2020long}, we use a gradually increasing parameter $\tau$ to adjust the logit. As we find in the ablative analysis that a proper number of $\tau$ may lead to a more balanced feature model, we attempt to use a smooth function to extend the experiment on $\tau=0.5$. To be specific, we replace the parmeter $\tau$ in Eq.~\ref{logit-adj loss} to $\alpha$ which will increase smoothly from zero to one as the training epochs increasing as follows:
\begin{align}
  \alpha = s  ({c^{ep}-1}),\quad s.t.\ ep = \frac{T}{T_{max}}\label{alpha},
\end{align}

\begin{align}
  \emph{loss} (y,z) & = -{\rm log} \frac{e^{z_y+\alpha \cdot {\rm log}\,{p_y}}}{\sum_{y'}^{L} e^{z_{y'}+\alpha \cdot {\rm log}\,{p_{y'}}}},
\end{align}
where $s$ and $c$  are scaling parameters, $T$ is the current epoch and $T_{max}$ is the number of total training epochs. With different $c$, $\alpha$ will be in concave, linear or convex forms. In practice, we find that $\alpha$ with convex form  performs better in our experiment. 

Intuitively, we design the adapting strategy for $\alpha$ based on the finding in ablative analysis. The change of $\alpha$ forces the deep model to fit the tail classes and increase the margin of tail classes gradually. From another side, the implement of $\alpha$ makes the network first learn the easy knowledge from the head classes sufficiently and gradually focus more on the hard knowledge from the tail classes which have few samples. 

\subsection{Adaptive Feature Generator}
Given the trained feature model from stage one, we augment the tail classes in feature space using AFG module. An ideal feature generator should generate features capturing intra-class information and do not deviate much from the true manifold which hinders learning. To this end, we apply Tukey's Ladder of Powers transformation \cite{tukey1977exploratory} to the tail class samples before generation and assume Gaussian prior. This transformation changes the shape of a skewed distribution and makes it close to Gaussian distribution.

In the first step, we calculate the statistics of different classes including mean and covariance:

\begin{gather}
  \mu_k = \frac{1}{N_k} \sum_{i\in D_k}{\widetilde{x_i}}  , \\
  \Sigma_k = \frac{1}{N_k-1}\sum_{i\in D_k}( {\widetilde{x_i}-\mu_k} )( {\widetilde{x_i}-\mu_k} )^T,
\end{gather}
 where $k\in \{1,2,...L\}$, $\mu_k $ and $\Sigma_k$ denote the mean and covariance of the Gaussian distribution for class $k$ separately and $D_k$ denotes the sub-set of long-tail dataset $D$ belonging to class $k$.
 
To generate the feature samples for tail classes, we calibrate the distribution for tail class samples after Tukey's Ladder of Powers transformation. For each transformed feature $\widehat{x_i}$ belonging to tail classes, we compute the distance between $\widehat{x_i}$ and other classes $k$ which have more training samples as $d_{ik} = || \widehat{x_i}- \mu_k||$. Then, we choose $K$ classes with smallest distance as a support set $S$ for tail class feature $\widehat{x_i}$ while $K$ is set to 2 or 3 in the experiment.

\begin{gather}
    S = \{j|j \in \{ y_i\} \ and \ d_{ij}\in topK[ -d_{ik}]  \}.
\end{gather}

Given the support set $S$ for tail class sample $\widehat{x_i}$, we calibrate the distribution of feature $\widehat{x_i}$ as:

\begin{gather}
  \mu_{\widehat{x_i}} = (1-\beta)  {\widehat{x_i}} + \beta\frac{\sum_{j \in S} \omega_j \mu_j}{\sum_{j \in S} \omega_j} , \label{calib_mu} \\
  \Sigma_{\widehat{x_i}} = (1-\beta)^2 \Sigma_i + \beta^2 \frac{\sum_{j \in S} \omega_j \Sigma_j}{\sum_{j \in S} \omega_j} + \gamma ,\\
  \omega_j = \frac{N_j}{d_{ij}},\ j \in S,
\end{gather}
where $\gamma$ is an optional hyper-parameter to increase the degree of dispersion and $\beta$ is the adaptive confidence score of the calibration. Recall the insight behind adapting the calibration strategy is that: if AFG improves the performance of the tail classes, then we should be more confident to transfer the knowledge from the head classes and increase $\beta$; if we observe worse performance from AFG, that means the calibration strategy need to be less confident and $\beta$ will be decreased. For the parameter $\omega$, we want the tail class samples learn the distribution knowledge more from the classes with more training samples and closer to the tail class samples.

For each tail class sample, we calibrate the distribution like above and generate samples within the calibrated distribution. The sampling number for each tail class will depend on the training sample number gap between current tail class and the head class. For the head classes, we sample from the original distribution to balance the number of samples belonging to different classes in the dataset. In the experiment, head classes always witness a performance drop because they can not learn knowledge from other classes and the augmentation on tail classes push the decision boundary much away from tail classes. To alleviate the problem, we adopt knowledge distillation loss for head classes between the stage one model and stage two model. It can be regarded as a regularization on the augmentation, and we want the model first learns to recover the real distribution of tail classes, then tries to calibrate the deviated decision boundary for the head classes. The new loss for stage two presents as

\begin{align}
  loss(y,z) = loss_{CE}(y,z)+\alpha \ loss_{KD}(M_1,M_2;z),\label{stage2 loss}
\end{align}
where $loss_{CE}$ is the cross-entropy loss, and $M_1,M_2$ is the model from stage one and stage two respectively. The $\alpha$ shares the same form as Eq.~\ref{alpha}. We want to first expand the decision boundary away from tail classes and generate tail samples adaptively. Then, in the latter epochs of stage two, the learning rate is much smaller and we want to push back the decision boundary based on the knowledge from stage one. In stage one, the model for head classes are trained sufficiently and it can be regarded as a teacher to guide the stage two training.

\section{Experiments}
\subsection{Experiment Settings}
\subsubsection{Datasets}  \ We evaluate our proposed method on three benchmark long-tail classification datasets: CIFAR100-LT \cite{cao2019learning}, ImageNet-LT \cite{kang2019decoupling} and iNaturalist \cite{van2018inaturalist}. The imbalance ratio of a dataset is defined as the training instances in the largest class divided by that of the smallest as $IM = \frac{max(N_i)}{min(N_i)}$. CIFAR100-LT is created with exponential decay imbalance with different imbalance ratio. ImagNet-LT is created by sampling a long-tail subset of full ImagNet training set following the Pareto distribution . It contains 1000 classes and the maximum class size is up to 1,280 while the minimum class size is 5 ($IM = 256$). Different from the hand-craft long-tail dataset, iNaturalist is a real-world fine-grained species classification dataset. Its 2018 version contains 437,513 training samples in 8,142 classes and the maximum class size is up to 1,000 while the minimum class size is 2  ($IM = 500$).

\vspace{-5mm}

\subsubsection{Implementation Details}  \ In our experiments, we use SGD optimizer with momentum 0.9 and weight decay 0.0005 for training. For CIFAR100-LT, we use ResNet-32 as backbone like \cite{cao2019learning}. Following the augmentation in \cite{ren2020balanced}, we train the model with cosine scheduler in stage one and train the classifier for 40 generation epochs. For ImageNet-LT, we first train a model without any bells and whistles following \cite{kang2019decoupling} using ResNet-10 and ResNet-50 as backbone. The model is trained for 200 epochs in the stage one and 40 generation epochs in the stage two with cosine scheduler. Similarly, we train the model on iNaturalist with pre-trained SSP model \cite{yang2020rethinking} for 90 epochs in the stage one and 40 generation epochs in the stage two. SSP \cite{yang2020rethinking} is a self-supervised pre-training method on iNaturalist. In the experiments, $\alpha$ in Gradual Balanced Loss is set with $c=2$, and $s=1$ for CIFAR100-LT and ImageNet-LT and $s=0.5$ for iNaturalist. The adaptive confidence score $\beta$ is initialized to 0.6 for tail classes and $\gamma$ is selected from \{0.0, 0.1\}. The $\alpha$ in Eq.~\ref{stage2 loss} is set the same as that in Gradual Balanced Loss.

\subsection{Experiment Results}

\subsubsection{Long-tailed CIFAR} \; Four datasets of different imbalance ratio, \{10, 50, 100, 200\}, are created for CIFAR-100. The results are shown in Table~\ref{table:CIFAR100}, we can see that the proposed method GLAG achieves SOTA performance compared with all the baselines. Comparing with the best re-weighting method Balanced Softmax \cite{ren2020balanced}, GLAG shows improvement by around 1\% on CIFAR100-LT with different imbalance ratio.
For data augmentation methods, GLAG also performs better than the SOTA approach SSD \cite{li2021self} and MetaSAug-LDAM \cite{Li_2021_CVPR}. 

\vspace{-5mm}

\subsubsection{Long-tailed ImageNet} \;  We also evaluate our proposed method on the large-scale long-tail dataset ImageNet-LT. We use ResNet-10 and ResNet-50 as our backbone. GLAG achieves a higher performance score which is higher than the SOTA method by at lest 0.4\% as shown in Table \ref{table:ImageNet result}. We find that GLAG benefits from larger epochs and deeper network.

\setlength{\tabcolsep}{4pt}
\begin{table}
\begin{center}
\caption{Top-1 classification accuracy on CIFAR100-LT}
\label{table:CIFAR100}
\setlength{\tabcolsep}{4mm}{
\begin{tabular}{c|cccc}
    \toprule
    Dataset& \multicolumn{4}{c}{CIFAR100-LT}  \\
	\hline  
	IM & 200 & 100 & 50 & 10  \\
	\hline
	End-to-end training& & & &\\
	\hline
    Cross-Entropy  &35.6  & 39.8 & 44.9 & 55.8\\
    Focal Loss \cite{lin2017focal}  &35.6  & 38.4 & 44.3 & 55.8\\
    LDAM-DRW \cite{cao2019learning} & 38.5 & 42.0 & 46.6 &  58.7\\
    Logit Adjust \cite{menon2020long}  & - & 43.89 & - &  -\\
    Balanced Softmax \cite{ren2020balanced}  & 45.5 & 50.8 & 54.5 & 63.0\\
    \hline
	Decoupled training& & & &\\
	\hline
    Decouple \cite{kang2019decoupling}  &38.2 & 43.3 & 47.4 & 57.9\\
    BBN \cite{zhou2020bbn} & -  &  42.6 & 47.1 & 59.1\\
    Bag of tricks \cite{zhang2021bag}  &-  & 47.8 & 51.7 & -\\
    MiSLAS \cite{zhang2021bag}  &-  & 47.0 & 52.3 & 63.2\\
    \hline
	Data augmentation& & & &\\
	\hline
	Remix-DRW \cite{chou2020remix} &-  & 46.8 & - & 61.3\\
	M2m \cite{kim2020m2m} &-  & 43.5 & - & 57.6\\
	RSG \cite{wang2021rsg} &-  & 44.6 & 48.5 & -\\
	MetaSAug-LDAM \cite{Li_2021_CVPR} & 43.1 & 48.1 & 52.3 & 61.3\\
	SSD \cite{li2021self} & - & 46.0 & 50.5& 62.3\\
    GLAG (ours) & {\bf46.9} & {\bf51.7} & {\bf55.3} & {\bf64.5} \\
    \bottomrule
\end{tabular}}
\end{center}
\end{table}
\setlength{\tabcolsep}{1.4pt}

\vspace{-20mm}
\begin{minipage}{\textwidth}
 \begin{minipage}[t]{0.55\textwidth}
  \centering
    \makeatletter\def\@captype{table}\makeatother\caption{Top-1 classification accuracy on ImageNet-LT}
    \vspace{3mm}
     \label{table:ImageNet result}
      \begin{tabular}{ccc} 
      \toprule
            Method & ResNet-10 & ResNet-50\\
            \hline
            Cross-Entropy  & 37.1 & 41.6 \\
            Focal Loss \cite{lin2017focal}     & -    & 43.7\\
            LDAM-DRW \cite{cao2019learning}    & 40.7 & 48.8\\
            Logit Adjust \cite{menon2020long}  & 43.0 & 51.1\\
            cRT \cite{kang2019decoupling}      & 41.8 & 47.3\\  
            LWS \cite{kang2019decoupling}      & 41.3 & 47.7\\
            OFA \cite{chu2020feature}          & 35.2 &-\\
            MetaSAug \cite{Li_2021_CVPR}       & -&50.5\\
            Bag of tricks \cite{zhang2021bag}  &43.1 &-\\
            GLAG(ours)  & {\bf43.3}  &{\bf51.5}\\
        \bottomrule
	\end{tabular}
  \end{minipage}
  \hspace{2mm}
  \begin{minipage}[t]{0.3\textwidth}
  \centering
    \makeatletter\def\@captype{table}\makeatother\caption{Top-1 accuracy on iNaturalist18}
    \vspace{3mm}
    \label{table:iNat result}
         \begin{tabular}{cc}        
         \toprule
            Method & ResNet-50\\
            \hline
            Cross-Entropy  & 61.7 \\
            LDAM-DRW \cite{cao2019learning}    & 64.6\\
            Logit Adjust\cite{menon2020long}  & 68.4\\
            BBN \cite{zhou2020bbn}             & 66.3 \\
            cRT \cite{kang2019decoupling}      & 65.2\\  
            LWS \cite{kang2019decoupling}      & 65.9\\
            OFA \cite{chu2020feature}          & 65.9\\
            MetaSAug \cite{Li_2021_CVPR}       & 68.7\\
            RSG \cite{wang2021rsg}             &67.9\\
            GLAG(ours)  & {\bf69.2}\\
        \bottomrule
	  \end{tabular}
  \end{minipage}
\end{minipage}

\subsubsection{iNaturalist} \;  iNaturalist is a real-world classification dataset containing 8,142 fine-grained species with imbalance ratio 500. We use ResNet-50 as our backbone following \cite{wang2021rsg}. As shown in Table \ref{table:iNat result}, GLAG outperforms baseline methods by at least 0.3\% and GLAG is higher than the SOTA augmentation method, RSG \cite{wang2021rsg}, by 1\%. The result shows that GLAG is capable of handling large-scale and highly imbalanced dataset.

\subsection{Ablation Study}

\subsubsection{$\alpha$ in Gradual Balanced Loss} \  One major hyper-parameter in our method is $\alpha$ in Eq.~\ref{alpha}, it adjusts the degree of adjustment in Gradual Balanced Loss. To be specific, the form of $\alpha$ can be divided to three types: linear, concave and convex as shown in Fig.~\ref{fig:alpha}. In Table \ref{table:alpha ablation}, we observe that $\alpha$ in convex form with $c=2$ works best and the score is higher to that of cross-entropy. Comparing to other forms, $\alpha$ in convex form with $c=2$ increases gently, which is more fit to the cosine scheduler. When the learning rate is large during training, $\alpha$ changes slowly and when the learning rate is small in latter training epochs, $\alpha$ is much higher and changes fast to help with the corresponding classifier balancing and increase the margin of tail classes.

\begin{minipage}{\textwidth}
  \centering
 \begin{minipage}{0.4\textwidth}
  \centering
    \makeatletter\def\@captype{table}\makeatother\caption{Ablation study on different form of $\alpha$ with the same setting as ablative analysis of feature model in Section 3.2.  }
    \vspace{3mm}
    \label{table:alpha ablation}
        \setlength{\tabcolsep}{4mm}{
        \begin{tabular}{cc}
            \toprule
            form of $\alpha$ & Acc\\
            \noalign{\smallskip}
            \hline
            \noalign{\smallskip}
                        linear & 57.26\\
            concave  & 56.73\\
            convex$(c=4)$ & 56.96\\
            convex$(c=6)$ & 56.56\\
            convex$(c=8)$ & 57.06\\
            convex$(c=2)$ & 57.69\\
            \bottomrule
        \end{tabular}}
  \end{minipage}
  \hspace{1mm}
  \begin{minipage}[h]{0.4\textwidth}
    \includegraphics[width=0.9\textwidth]{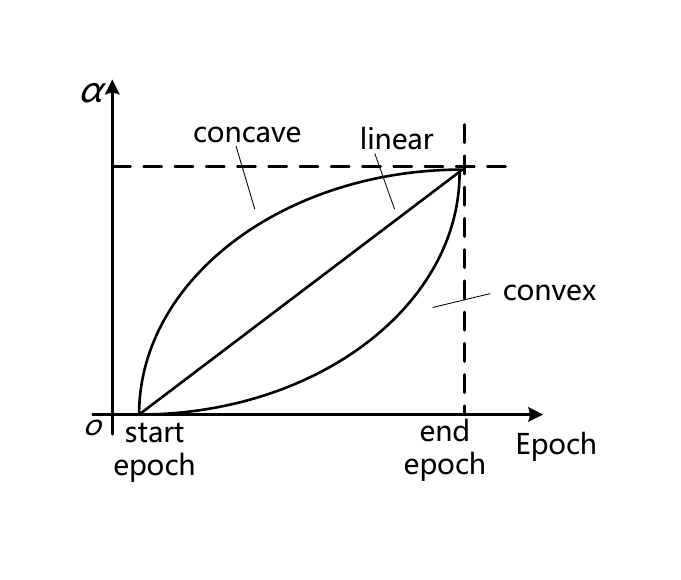}
    \makeatletter\def\@captype{figure}\makeatother\caption{Different form of $\alpha$ in Eq.~\ref{alpha}}
    \label{fig:alpha}
  \end{minipage}
\end{minipage}

\subsubsection{$\beta$ in Adaptive Feature Generator} \ Recall in Eq.~\ref{calib_mu}, we introduce a class-wise adaptive confidence score $\beta$ which control the degree of distribution calibration. We initialize $\beta$ to 0.4 for each tail class and it changes  adaptively during training. Overall, the tail classes with rare samples exhibit higher confidence score comparing with the tail classes with few samples. As we explained, if AFG improves the performance of the tail classes, then we should be more confident to transfer the knowledge from the head classes and increase $\beta$. For the tail classes with rare samples, the increase on $\beta$ shows performance enhancement during training. For the tail classes with few samples, they have more samples and do not need to learn much knowledge from other classes, so the confidence score is lower.




\subsubsection{Analysis on feature learning in stage one} 
\begin{figure}
\centering
\includegraphics[scale=0.55]{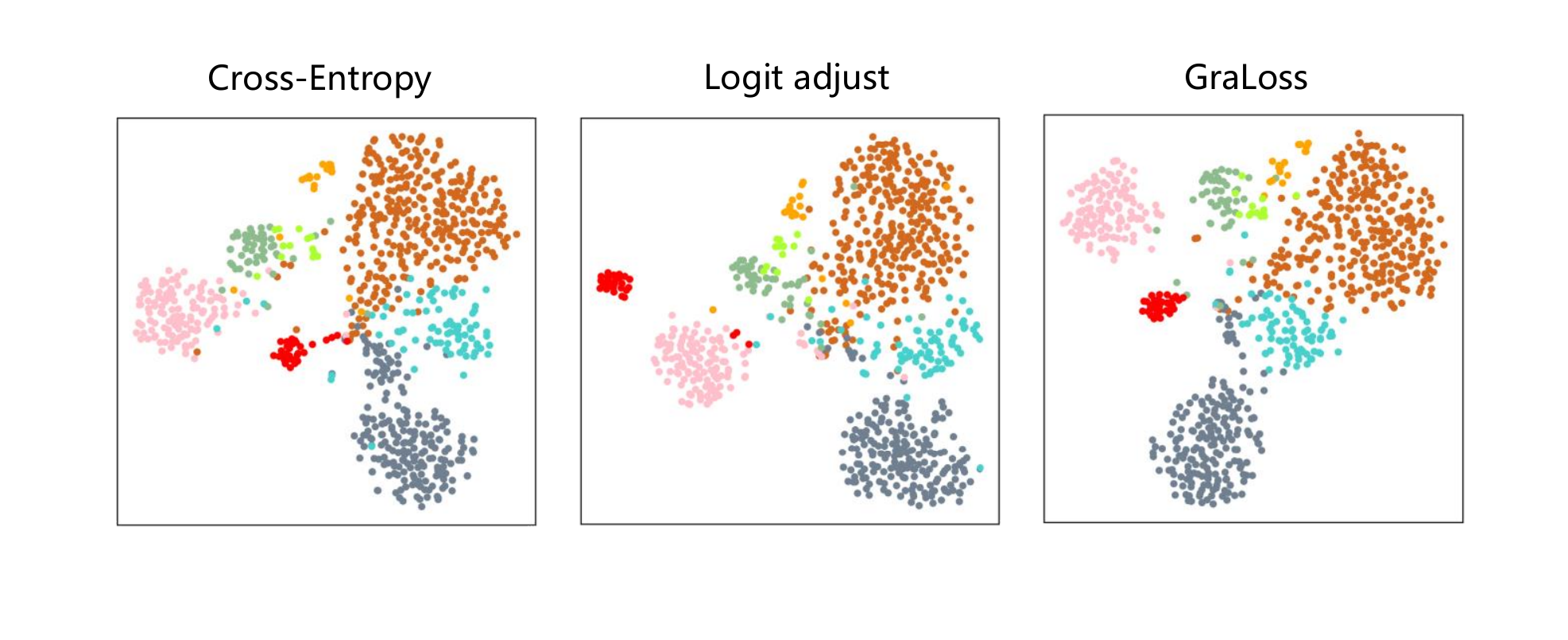}
\vspace{-10mm}
\caption{Visualization of different loss function guided feature learning in stage one.}
\label{fig:tsne}
\end{figure}

\ In Sec.~\ref{Ablative Analysis}, we have analyzed the effects of different loss function to feature model in stage one. In this subsection, we visualize the feature belonging to eight classes which we selected randomly from long-tail CIFAR-100 dataset via t-SNE. As shown in Fig.~\ref{fig:tsne}, features trained with GraLoss are more separable than logit adjust \cite{menon2020long} and cross-entropy. For example, when training with GraLoss, the light-blue class samples cover smaller space and can be well separated from others. Instead, when training with logit adjust and cross-entropy, some light-blue class samples are mixed with other classes.

Furthermore, we want to claim the generality of our GraLoss for different decoupled training methods. We select three SOTA methods: cRT \cite{kang2019decoupling}, LWS \cite{kang2019decoupling} and MiSLAS \cite{zhong2021improving}. All these methods are trained with cross-entropy in stage one and we simply replace the original cross-entropy to GraLoss and re-weighting loss LogitLoss \cite{menon2020long}. As Table~\ref{table:GraLoss&decouple} shown, for cRT and MiSLAS, performance of cross-entropy is higher than LogitLoss and similar performance drop is also claimed in \cite{ren2020balanced}. For all three decoupled training methods, using GraLoss in stage one helps the model achieve higher performance by 0.2\%-0.8\%. These results demonstrate the high practicability for decoupled methods, one can simply combine GraLoss and other decoupled methods to achieve higher performanc.

\setlength{\tabcolsep}{4pt}
\begin{table}
\begin{center}
\caption{Ablation study on CIFAR100-LT for different decoupled methods. In stage one, we use different loss function to train the feature model. In stage two, we use different decoupled methods to train the classifier. }
\label{table:GraLoss&decouple}
\setlength{\tabcolsep}{4.5mm}{
\begin{tabular}{cc|cccc}
    \toprule
    \multicolumn{2}{c}{Dataset}& \multicolumn{4}{c}{CIFAR100-LT}  \\
	\hline  
	\multicolumn{2}{c}{IM} & 200 & 100 & 50 & 10  \\
	\hline
    \multirow{3}{*}{cRT \cite{kang2019decoupling}}   &  CE & 46.3  & 50.2  & 54.6  & 63.8\\
    &  GraLoss & 46.6  & 50.5  & 54.7  & 63.9\\
    &  LogitLoss & 46.2  & 50.0  & 54.5  & 63.6\\
    \hline
    \multirow{3}{*}{LWS \cite{kang2019decoupling}}   &  CE & 44.9  & 50.4  & 54.1  & 63.5\\
    &  GraLoss & 46.4  & 50.6  & 54.9  & 64.1\\
    &  LogitLoss & 46.1  & 50.4  & 54.2  & 63.3\\
    \hline
    \multirow{3}{*}{MiSLAS \cite{zhong2021improving}}   &  CE & 45.7  & 49.4  & 53.5  & 62.2\\
    &  GraLoss & 46.2  & 49.8  & 53.6  & 62.4\\
    &  LogitLoss & 45.3  & 49.2  & 53.3  & 62.2\\
    \bottomrule
\end{tabular}}
\end{center}
\end{table}
\setlength{\tabcolsep}{1.4pt}

\vspace{-10mm}
\subsubsection{Effectiveness of each components} \ We investigate the contribution of each components in our proposed method and compare AFG to combination with different loss function in stage one, which is shown in Table \ref{table:GLAG abltaion}. We observe AFG combining with cross-entropy performs better than that with logit adjust loss. The result reflects from another point of our ablative analysis. A balanced and robust feature model indeed benefits the classifier tuning, so the method combining with GraLoss can achieve the best.

Comparing to the results in stage one, AFG with adaptive confidence score $\beta$ has better performance. Although we can set $\beta$ to a certain number which can have similar results, our method adjusts $\beta$ adaptively on class-wise level. It is apparent in Eq.~\ref{calib_mu} that the confidence score belonging to different classes varies much. The adaptive method avoids expensive hyper-parameter tuning on the confidence score, which makes AFG easier to apply on different dataset. 

Knowledge distillation loss is another important component in our method. As shown in Table \ref{table:GLAG abltaion}, after the tuning with knowledge distillation loss on head classes, the performance of tail classes drops while the overall performance increases. That is like a seesaw: when the learning rate is large, we generate tail class samples to push the boundary away from tail classes, and when the learning rate is small, we push back the boundary to balance the performance on different classes.

\setlength{\tabcolsep}{4pt}
\begin{table}
\begin{center}
\caption{Ablation study of GLAG on CIFAR100-LT. CE: cross-entropy loss in stage one. GraLoss: gradual balanced loss in stage one. Logitloss: logit adjust loss \cite{menon2020long} in stage one. AD: 
adaptive confidence score $\beta$ in stage two. KD: knowledge distillation loss for head classes in stage two.}
\label{table:GLAG abltaion}
\setlength{\tabcolsep}{2.2mm}{
\begin{tabular}{ccccc|cccc}
    \toprule
    CE  &  GraLoss    & Logitloss  & AD  & KD & Average & Many & Medium & Few\\
    \noalign{\smallskip}
    \hline
    \noalign{\smallskip}
    \Checkmark   & \XSolidBrush & \XSolidBrush &  \XSolidBrush  &  \XSolidBrush  & 45.3 & 72.5 & 45.8 & 13.1 \\
    \XSolidBrush & \Checkmark   & \XSolidBrush &  \XSolidBrush  &  \XSolidBrush  & 50.4 & 70.1 & 49.0 & 29.0\\
    \XSolidBrush & \XSolidBrush & \Checkmark   &  \XSolidBrush  &  \XSolidBrush  & 50.3 & 67.7 & 50.2 & 30.3\\
    \XSolidBrush & \Checkmark   & \XSolidBrush &  \Checkmark    &  \XSolidBrush  & 51.4 & 64.3 & 53.3 & 34.4\\
    \XSolidBrush & \XSolidBrush & \Checkmark   &  \Checkmark    &  \Checkmark    & 50.6 & 64.6 & 51.3 & 33.6\\
    \Checkmark   & \XSolidBrush & \XSolidBrush &  \Checkmark    &  \Checkmark    & 50.8 & 63.7 & 53.0 & 33.2\\
    \XSolidBrush & \Checkmark   & \XSolidBrush &  \Checkmark    &  \Checkmark    & {\bf51.7} & 65.8 & 52.9 & 33.9\\

    \bottomrule
\end{tabular}}
\end{center}
\end{table}
\setlength{\tabcolsep}{1.4pt}


\vspace{-10mm}
\section{Conclusions}

In this paper, we propose a new method GLAG to tackle the long-tail challenge. We introduce two components in decoupled training to improve the performance. The first component Gradual Balanced Loss is a general loss for different decoupled training method which is different from previous re-weighting loss. In ablative analysis, we find that existing re-weighting loss can not help with decoupled training. However, our method can improve the performance of feature model and outperforms the widely adopted cross-entropy method in decoupled training. The second component feature generator generates tail class samples adaptively. Different from previous augmentation methods, we introduce distribution learned augmentation method and knowledge distillation loss on head classes to balance the performance. It can be regarded as the regularization to the feature generator.




\clearpage
%
%
\bibliographystyle{splncs04}
\bibliography{egbib}
\end{document}